\documentclass[conference]{IEEEtran}
\IEEEoverridecommandlockouts

\usepackage{amsmath,amssymb,amsfonts} % Math env and fonts
\usepackage{hyperref} % Links
\usepackage{graphicx} % Include graphics
\usepackage{booktabs,dsfont}
\usepackage{bbold}
\usepackage{cite}
\usepackage{algorithmic}
\usepackage{textcomp}
\usepackage{xcolor}
\def\BibTeX{{\rm B\kern-.05em{\sc i\kern-.025em b}\kern-.08em
T\kern-.1667em\lower.7ex\hbox{E}\kern-.125emX}}

\newcommand{\encoding}{\mathsf{z}^e}
\newcommand{\quantized}{\mathsf{z}}
\newcommand{\codebook}{\mathrm{e}}
\newcommand{\argmin}{\mathtt{argmin}}
\newcommand{\argmax}{\mathtt{argmax}}
\newcommand{\elbo}{\mathcal{L}}

\newcommand{\expectation}{\mathbb{E}_{q_\varphi}}
\newcommand{\command}{u}
\newcommand{\softmax}{\mathtt{softmax}}

\def\BibTeX{{\rm B\kern-.05em{\sc i\kern-.025em b}\kern-.08em
T\kern-.1667em\lower.7ex\hbox{E}\kern-.125emX}}

\title{Variational Latent Discrete Representation for Time Series Modelling
\thanks{This work was supported by grants from Région Ile-de-France.}
}

\author{\IEEEauthorblockN{Max Cohen}
	\IEEEauthorblockA{\textit{T\'el\'ecom SudParis, CITI, TIPIC} \\
		Institut Polyechnique de Paris \\
		maxjcohen@proton.me}
	\and
	\IEEEauthorblockN{Maurice Charbit}
	\IEEEauthorblockA{\textit{Accenta, Boulogne-Billancourt} \\
		maurice.charbit@accenta.fr}
	\and
	\IEEEauthorblockN{Sylvain Le Corff}
	\IEEEauthorblockA{\textit{T\'el\'ecom SudParis, CITI, TIPIC} \\
		Institut Polyechnique de Paris}
}

\begin{document}
\maketitle

\begin{abstract}
	Discrete latent space models have recently achieved performance on par with their continuous counterparts in deep variational inference.
	While they still face various implementation challenges, these models offer the opportunity for a better interpretation of latent spaces, as well as a more direct representation of naturally discrete phenomena. Most recent approaches propose to train separately very high-dimensional prior models on the discrete latent data which is a challenging task on its own.
	In this paper, we introduce a latent data model where the discrete state is a Markov chain, which allows fast end-to-end training.
	The performance of our generative model is assessed on a building management dataset and on the publicly available Electricity Transformer Dataset.
\end{abstract}

\begin{IEEEkeywords}
	Discrete latent space, Recurrent Neural Networks, Variational Inference, ELBO.
\end{IEEEkeywords}

\section{Introduction}
Discrete latent space models aim at representing data through a finite set of features.
Recent advances in generative models have pushed towards these representations, as they fit data with naturally discrete hidden states.
For instance, defining a classification task over a dataset often implies a partition of an unobserved latent space, which could be modelled using a categorical random variable, as presented in \cite{Kingma2014}.
In attention-based models, representing the focus location as a discrete variable, i.e. the right place to focus in the past for predicting the next observation, has proven efficient and can help interpreting prediction errors, see for instance \cite{Xu2015}.
When analyzing time series data, the evolution of a discrete latent variable can be interpreted as a switch in regime, see for instance \cite{Dzunic2014,lindsten2015rao,nguyen2017particle}.
In \cite{Ajib2020}, the authors model the indoor temperature of a building by identifying discrete regimes, such as the opening of a window, the presence of occupants or shade.

As one of the most widespread expression of discrete latent space models, Hidden Markov Models have been successfully applied in various fields (\cite{Wilks1998, Gales2008, Patterson2017}).
Despite modeling hidden states as a discrete Markov chain, they are able to handle complex data structure, see \cite{Cappe2005,douc2014nonlinear} and the references therein for a complete overview.
However, dealing with latent data leads to models that are computationally expensive to train, for instance using Expectation Maximization \cite{Dempster77EM} based approaches, and still struggle to handle large scale datasets, in particular when the models contain high-dimensional additional latent states.

In contrast, deep learning methods are able to infer millions of parameters from huge amounts of data - at the cost of much more complex models - through automatic differentiation and gradient computation.
However, discrete variables usually prevent the propagation of the gradient, and in turn straightforward trainings.
The Vector-Quantized Variational AutoEncoder (VQ-VAE, \cite{Oord2017}) popularized discrete latent spaces for Variational Inference (see \cite{Jordan2004AnIT} for an introduction), by copying the gradient of discrete variables directly to the previous layer.
This approximation is detailed in \cite{Bengio2013EstimatingOP}.
Additionally, the training of the VQ-VAE is divided into two steps.
The first assumes a non informative prior on the latent variables in order to estimate the encoder and decoder parameters.
Samples generated after this step cannot catch the complex dependencies that lie in the latent space; to produce coherent samples, a prior model on the latent variables is trained separately, directly on the latent space.
Because these prior models are usually already challenging to fit on their own, most applications of the VQ-VAE have not been able to jointly train both parts of the network, see \cite{Sun2020}.
This paper brings the following contributions.
\begin{itemize}
	\item We propose a generative model for time series, where the latent space is modelled as a discrete Markov chain, and, conditionally on the latent states, the observations follow a simple autoregressive process. Parameters are jointly estimated by Variational Inference.
	\item We compared the impact of different prior models to extract high-level features from the input data (based on convolutional and recurrent architectures) on the quality of samples.
	\item Our model outperforms the state of the art VQ-VAE in both accuracy and computation time on two time series datasets.
\end{itemize}

\section{Related works}
\paragraph{Discrete latent representation}
one of the most straightforward application of discrete latent representation is derived from semi or unsupervised classification problems, where the data presents distinct semantic classes.
The authors of \cite{Kingma2014} propose to model such data as generated by both a continuous and a categorical class variable.
By integrating a classification mechanism directly in the model, they are able to outperform continuous latent models, while allowing conditional generation.
This idea was transposed to Generative Adversarial Networks (introduced in \cite{Goodfellow2014GAN}) by the authors of \cite{Chen2016}, by adding a discrete random variable as the input of the generator.
Trainings on MNIST show how the model associates each value with a class of digits, even without supervision.

The interest in quantized latent variables go beyond classification tasks, and appear in attention-based models, which have dominated the Natural Language Processing and Computer Vision fields for the last years.
For instance, \cite{Xu2015} presents an image captioning model where attention scores parametrize a categorical random variable; by either sampling or computing its expectation, the model extracts visual features used to generate words.
Discrete representations have also successfully replaced continuous latent spaces in existing models.
In this way, the authors of \cite{Sun2020} quantized the latent features of an existing variational recurrent neural network applied to text-to-speech synthesis, and reported improved quality in generated audio samples.

\paragraph{Variational Inference}
Variational Inference (VI) is a popular approach to estimate such models, which consists in approximating the posterior distribution of the latent states given the observations, within a parametric family of tractable distributions, see \cite{Salakhutdinov2009, Mnih2016,petetin2021structured}.
A trade-off between the expressiveness of the approximated posterior, and its computation cost must be achieved, which represents the main challenge of VI methods.
As the computation of the likelihood of the model and its gradient are intractable, the training procedure usually focuses on maximizing a lower bound of the likelihood by gradient ascent.
As sampling discrete random variables prevent gradient propagation through the model, a workaround is required to take advantage of automatic differentiation offered by modern deep learning frameworks.
The usual alternative to Variational Inference for modelling time dependant data is Sequential Monte Carlo, see for instance \cite{Cohen2023LastLayerDecoupling} and the references therein.

Vector Quantized Variational Auto Encoders introduced a two-stage training for discrete latent models: first, the posterior is approximated through VI while considering an uninformative prior for the discrete latent variables, i.e. a uniform discrete distribution; then, a very expressive prior model is trained on the discrete latent space to produce samples, all other parameters being fixed.
Despite its impressive results, the VQ-VAE still suffers from two drawbacks: the differentiation of the posterior and the choice of the prior.

\paragraph{Differentiation of the posterior}
sampling discrete variables from the posterior prevents gradient propagation through the model; while the authors of the VQ-VAE proposed a straight-through estimator, other approaches have addressed this issue.
In \cite{Lorberbom2019}, the model is optimized through direct loss minimization, a method that introduces additional bias and hyperparameter tuning.
Another method, introduced in \cite{Bartler2019}, leverage importance sampling to sample from the posterior, without introducing bias or new parameters.
However, a new differentiable distribution close to the posterior must be introduced, limiting its usage in general settings.
The authors of \cite{Jang2017CategoricalRW} proposed a new differentiable distribution, approximating samples from a categorical law while being differentiable.
Although this method does introduce a new hyperparameter, it offers a very appealing trade-off.

\paragraph{Prior}
during the second stage of the VQ-VAE training, an autoregressive prior is trained on the latent discrete space.
In the original paper, the authors chose very resource intensive models for both image \cite{Oord2016Pixelcnn} and audio \cite{Oord2018WaveNet} datasets.
However, more parsimonious prior models already yield encouraging results.
For instance, we can find in \cite{Sun2020} a comparison of different autoregressive priors on a text-to-speech task, leading to state of the art performances, using a single layered Long Short Term Memory (LSTM) network.
On the same task, the authors of \cite{Yasuda2021} were able to perform end-to-end training.
Additionally, this prior model was conditioned on text data to produce speech samples.

\section{Model}
Consider a sequence of observations $(x_1,\ldots,x_T)$ in $\mathbb{R}^d$.
It is assumed that these observations depend on external observed signals $(u_{1},\ldots,u_T)$ in $\mathbb{R}^q$, referred to as commands.
In this paper, we also assume that the observations are independent of the commands conditionally on hidden variables $(\quantized_1,\ldots,\quantized_T)$, which take values in a set of codebooks $\mathcal{E} = \{\codebook_1,\ldots,\codebook_K\}$.
For each of $1\leq k\leq K$, $\codebook_k\in\mathbb{R}^D$.
We then consider the following family of probability density functions:
\begin{equation}
	\label{eq:model}
	p_\theta(y_{1:T} | u_{1:T}) = \int p_\theta(y_{1:T} | \quantized_{1:T}) p_\theta(\quantized_{1:T} | u_{1:T}) d\quantized_{1:T}\,,
\end{equation}
depending on an unknown parameter $\theta\in\Theta\subset \mathbb{R}^m$ and where for all $1\leq s\leq t$ and all sequence $\{a_u\}_{u\geq 1}$, $a_{s:t}$ is a short-hand notation for $(a_s,\ldots,a_t)$.

Conditionally on the commands, we assume that the latent states are Markovian and that the conditional law of the observations depends on past latent states, so that our model is more general than Hidden Markov Models.
It also differs from other extensions of HMMs with dependencies between the observations, such as the autoregressive processes as described in \cite{Douc2004}.
In the following paragraphs, we detail the structure of the observation and prior models, as well as the choice of posterior family.

\subsection{Observation model}
We consider a Gaussian observation model, and estimate at each time step its mean and variance:
$p_\theta(x_{1:T} | \quantized_{1:T}) = \prod_{t=1}^T \varphi_{\mu_t, \sigma_t^2} (x_t)$,
where $\varphi_{\mu, \sigma^2}$ is the probability density function of a Gaussian random variable with mean vector $\mu$ and covariance matrix $\sigma^2 I_d$.
For all $1 \leq t \leq T$, $\mu_t = g_\theta^\mu(\mu_{t-1}, \quantized_{1:t-1})$ and $\sigma_t = g_\theta^\sigma(\sigma_{t-1}, \quantized_{1:t-1})$, with $\mu_{0} \equiv \sigma_0 \equiv 0$.
We discuss our choice for the parametric functions $g_\theta^\mu$ and $g_\theta^\sigma$ in Section~\ref{sec:chosen_architectures}.

\subsection{Prior model}
We assume that, conditionally on the commands $\command_{1:T}$, the latent state is a discrete Markov chain.
Write, for all $1\leq \ell\leq K$, $p^{\ell}_{\theta,1} = p_\theta(\quantized_1 = \codebook_\ell | \command_{1:T})$, and for all $2\leq t\leq T$ and $1\leq k,j\leq K$, $p^{k,j}_{\theta,t|t-1} = p_\theta(\quantized_t = \codebook_k | \quantized_{t-1} = \codebook_j, \command_{1:T})$. The prior distribution is then defined as:
\begin{multline*}
	\log p_\theta(\quantized_{1:T} | \command_{1:T}) = \sum_{k=1}^K \mathds{1}_{\quantized_1 = \codebook_k} \log p^{k}_{\theta,1}\\
	+ \sum_{t=2}^T \sum_{j,k=1}^K \mathds{1}_{\quantized_{t-1} = \codebook_j}\mathds{1}_{\quantized_{t} = \codebook_k} \log p_{\theta, t|t-1}^{k,j}\,.
\end{multline*}

\subsection{Posterior distribution}
As the posterior distribution of $\quantized_{1:T}$ given $x_{1:T}$ is intractable, we use a variational approach to estimate $\theta$.
In the original approach proposed in \cite{Oord2017}, the authors use an encoding function $f_\varphi$, depending on an unknown parameter $\varphi \in\mathbb{R}^p$, mapping the observations to a series of encoded latent variables $\encoding_{1:T} = f_\varphi(x_{1:T})$.
Then, the posterior is approximated by $q_\varphi(\quantized_t = \codebook_k| x_{1:T}) = \mathds{1}_{\codebook_* = \codebook_k}$, where $\codebook_* = \argmin_{\codebook_\ell\in\mathcal{E}} \| \encoding_t - \codebook_\ell \|_2$.
We propose, for $1 \leq k \leq K$:
\begin{equation}
	\label{eq:posterior}
	q_{\varphi, t}^k = q_\varphi(\quantized_t = \codebook_k | x_{1:T})  \propto \exp\{ - \| \encoding_t - \codebook_k \|_2^2 \}\,.
\end{equation}
During both training and inference, this discrete distribution encourages the exploration of the entire set of codebooks. Under the variational distribution, the latent data are assumed to be independent conditionally on the observations.

\section{Inference procedure}
The Evidence Lower BOund (ELBO) can be decomposed in three terms: for all $(\theta,\varphi)$,
\begin{multline}
	\elbo(\theta,\varphi) = \expectation \left[\log p_\theta(x_{1:T} | \quantized_{1:T})\right]
	+ \expectation \left[\log p_\theta(\quantized_{1:T} | u_{1:T})\right]    \\
	-\expectation \left[\log q_\varphi(\quantized_{1:T} | x_{1:T}) \right]\,.
\end{multline}
The last term of the ELBO can be computed explicitly as follows:
$$
	\expectation [\log q_\varphi (\quantized_{1:T} | x_{1:T}) ] = \sum_{t=1}^T \sum_{k=1}^K q_{\varphi, t}^k \log q_{\varphi, t}^k\,.
$$
%The two other terms cannot be computed similarly as this would result in a huge computation overhead.
%Instead, we approximate this expectation with a single sample under the posterior.
% Check paper suedois sur le soft vqvae
%In order to optimize the negative ELBO using gradient descent algorithms,

We approximate the two other terms by drawing $M>0$ samples under $q_\phi$ and computing a Monte Carlo estimator.
In order for this operation to be differentiable, we use a reparametrization designed for categorical variables based on the Gumbel-Softmax distribution.

In \cite{Gumbel1955}, it is shown that we can draw samples under $q_\phi$ by computing $\argmax_{k=1}^K (\log(q_\phi^k) + g_k)$, where $(g_k)_{k=1}^K$ are independent identically distributed  samples from the Gumbel distribution $G(0, 1)$ with probability density $g(x) = e^{-(x+e^{-1})}$.
The Gumbel-Softmax distribution \cite{Jang2017CategoricalRW} aims at using the $\softmax$ function as a continuous and differentiable approximation to the $\argmax$ operator.
We sample $(g_1, \ldots, g_K)$ independently from the Gumbel distribution and define, for all $1 \leq k \leq K$, $\pi_{k,t} \propto \exp((\log q_{\phi,t}^k + g_k) / \tau_t)$, where $\tau_t>0$ is the softmax temperature, allowing for a smooth interpolation between a Uniform distribution (for large values of $\tau_t$), and a Categorical distribution (for small values of $\tau_t$).
We propose to approximate the variational posterior distribution of $\quantized_t$ by the Dirac mass at $\widetilde \quantized_t = \sum_{k=1}^K \pi_{k,t} \codebook_k$.
Through re-parametrization, this method allows for differentiation of the sampled latent vector $\widetilde \quantized_t$, with respect to the codebooks $\codebook_k$, $1\leq k \leq K$, and the encodings $ \encoding_{1:T}$.
The first term of the ELBO can now be approximated by $(\theta, \phi) \mapsto M^{-1} \sum_{i=1}^M \log p_\theta(\widetilde \quantized_{1:T} | u_{1:T})$, and the second by $(\theta, \phi) \mapsto M^{-1} \sum_{i=1}^M \log p_\theta(y_{1:T} | \widetilde \quantized_{1:T})$.

Estimating $\theta$, $\phi$ and $\mathcal E$ jointly can induce instability at the beginning of the training, leading to diminished performances after convergence.
In order to control the trade-off between the capacity of the latent state and the observation model, \cite{Higgins2017betavae} introduced an additional hyperparameter $\beta$ when computing the ELBO.
As shown in \cite{Ramesh2021}, we can perform end-to-end training by penalizing the prior and posterior terms by this factor $\beta$ initialized close to zero, and then slowly increasing its value until reaching $\beta = 1$.

\section{Experiments}
\begin{figure}[htpb]
	\centering
	\caption{We sample $N=100$ trajectories under the GRU prior, conditioned on a set of commands from the validation split of the ETT dataset, and plot the associated 95\% confidence intervals.
		As a baseline, we added samples from a VQ-VAE model.}
	\includegraphics[width=0.46\textwidth]{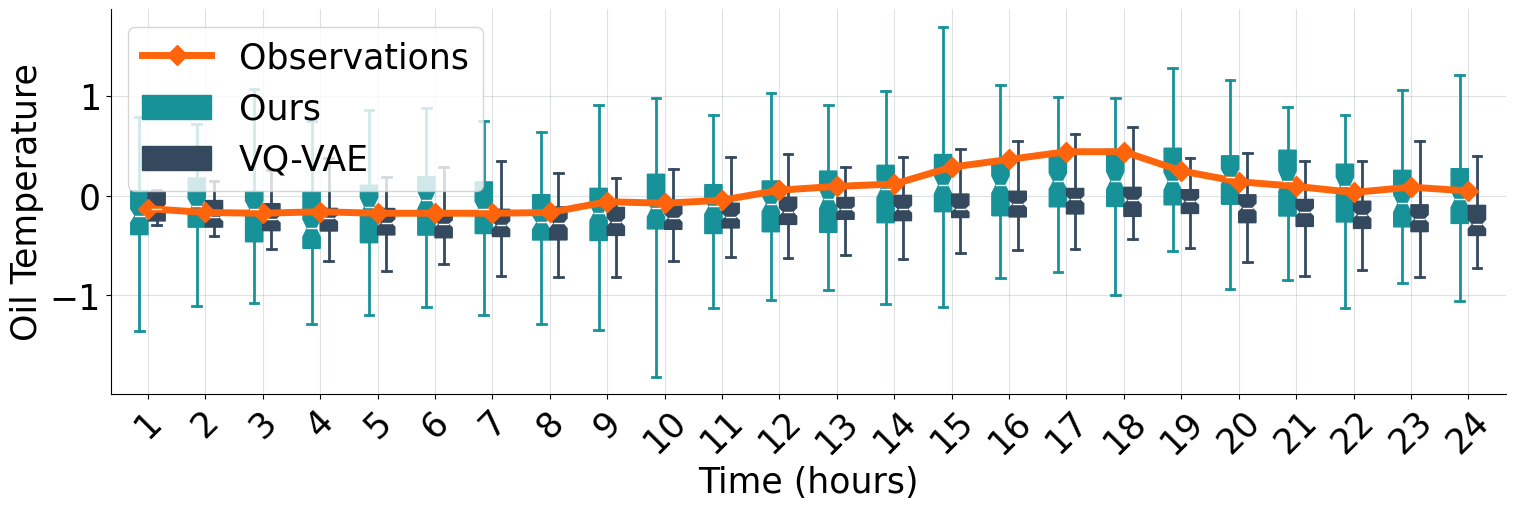}
	\includegraphics[width=0.46\textwidth]{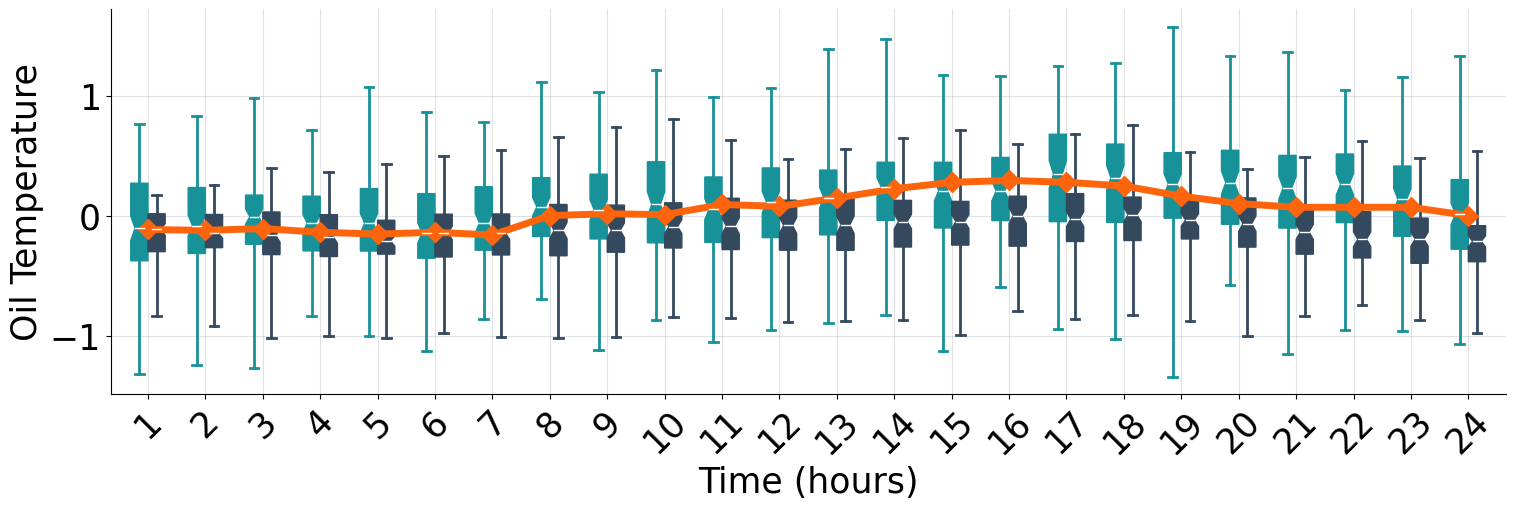}
	\label{fig:averaged-samples-comparison}
\end{figure}

We benchmarked our approach on the public Electricity Transformer Temperature (ETT) Dataset, designed in \cite{Zhou2021Informer} to forecast Oil temperature based on hourly power load records (ETTh1 subset).
Models are trained on the first year of available data, and validated on the subsequent four months.
Samples from the validation set are presented in Figure~\ref{fig:averaged-samples-comparison}, where we compare our proposed model with the benchmarked VQ-VAE.
We also propose a benchmark on a building management dataset, where we aim at forecasting hourly indoor humidity based on the indoor temperature and outdoor humidity.

\subsection{Chosen architectures}
\label{sec:chosen_architectures}
Our prior model can be decomposed in two sub-networks: an input model is responsible for extracting high level features from the commands, while the autoregressive kernel computes the Markov chain transition probabilities.
This disentanglement allows us to keep the kernel simple, while still working with high level features:
\begin{align*}
	\tilde u_{1:T} & = f_\theta^\text{input\_model} (u_{1:T}) \,,                                                   \\
	h_t            & = \ker_\theta (h_{t-1}, \tilde u_{t-\Delta:t}), \, \forall 1 \leq t \leq T, \, h_0 \equiv 0\,.
\end{align*}
For the input model, we implemented a 3-layered LSTM, with the same latent dimension as the commands.
We then compared several auto regressive architectures.
\begin{itemize}
	\item A simple RNN cell was used as a benchmark, as they struggle to model long term dependencies.
	\item A Gated Recurrent Unit (GRU) cell, following results in \cite{Sun2020}. These architectures have a more refined memory representation, and have shown encouraging results when applied to building management data, as shown in \cite{Cohen2021}.
	\item A kernel based on causal convolutions, as proposed in \cite{Oord2018WaveNet}, where memory is replaced by an explicit dependency on the last $\Delta=24$ time steps.
\end{itemize}
For the encoder $f_\varphi$ and decoder $g_\theta^{\mu, \sigma}$ parametric functions, we chose 3-layered LSTM networks as well.

\begin{figure}[htb]
	\centering
	\caption{Codebook usage for a week long sample of the ETT dataset.
		We sample $N=100$ trajectories from the prior, and plot the most selected codebook index at each time step.
		The observed oil temperature has been plotted for comparison.
		The second codebook seems to be correlated with an increase in oil temperature, while the third one is activated during a decrease.
		The seventh seems to only appear for low oil temperatures.}
	\includegraphics[width=0.43\textwidth]{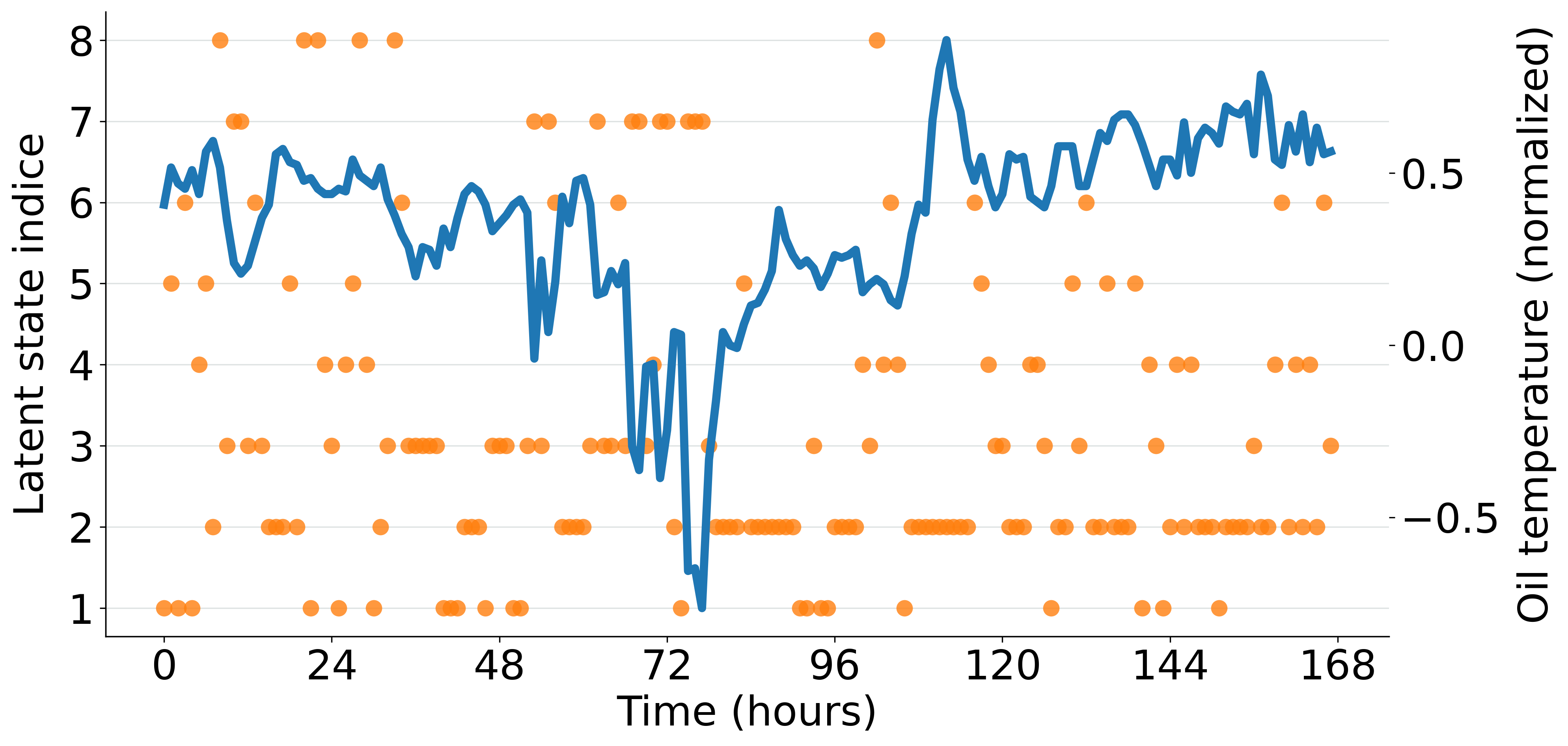}
	\label{fig:codebook_usage}
\end{figure}

Finally, we cross validated the number of codebooks using the Root-Mean-Square Error (RMSE), as described in Section~\ref{sec:evaluation}, and settled for $K=8$.
The influence of each codebook on the produced samples is illustrated in Figure~\ref{fig:codebook_usage}, where we plot the selected codebook during the generation process at each time step.
This representation allows for a deeper understanding of the latent representation of the model, while opening the way for further analysis such as regime switching detection.

We experimented with various schedules increasing $\beta$ from 0 to 1, and found that while this penalization is necessary to estimate all parameters jointly, the choice of schedule had little influence on the performances.
Therefore, we increase $\beta$ linearly between epochs 1 to 100, and keep $\beta=1$ for the rest of the training, which amounts to 300 total epochs.
In all following simulations, the codebooks dimension is fixed to $D=32$, and the number of Monte Carlo samples to $M=1$.

We compared our model to a Gaussian discrete linear Hidden Markov Model (HMM) whose parameters are estimated with the Kalman smoother using the Expectation Maximization (EM, \cite{Dempster77EM}) algorithm.
We used the same number of hidden states $K=8$.
We also compared our approach to the original VQ-VAE.
The architecture of the model is kept similar, to highlight the following difference of methodology and training.
\begin{itemize}
	\item The autoencoder is trained while considering an uninformative prior. Then, we estimate the parameters of the prior model while the autoencoder is fixed.
	\item Samples from the posterior distribution are drawn under a Dirac mass, as shown above. In order to compute the gradient, we use the straight-through estimator.
\end{itemize}

\subsection{Evaluation}
\label{sec:evaluation}
Provided a sequence of commands, each benchmark model produces samples predicting the observations.
By averaging them over the time steps, we compute the RMSE: let $\hat x_{1:T}^i, 1 \leq i \leq N$ be $N$ independent sequences of predictions of $x_{1:T}$, then $\text{RMSE} = (T^{-1} \sum_{t=1}^T (x_t - N^{-1} \sum_{i=1}^N \hat x_t^i)^2)^{1/2}$.
Additionally, we report the Mean Absolute Error (MAE) criteria: $\text{MAE} = T^{-1} \sum_{t=1}^T \|x_t - N^{-1} \sum_{i=1}^N \hat x_t^i\|$.
In the presented experiments, we set $N=100$.
%Let $\hat x_L^{1:T}, \hat x_U^{1:T}$ be the lower and upper bound of those samples, we also compute the Prediction Interval Coverage Probability (PICP) and the Mean Prediction Interval Width as $\mathrm{PICP} = T^{-1}\sum^{T}_{t=1} \mathbb{1}_{[\hat x_L^t, \hat x_U^t]}(x_t)$ and $\mathrm{MPIW} = T^{-1} \sum^{T}_{t=1}(\hat x_U^t - \hat x_L^t)$.
Results on the entire validation set are displayed in Table~\ref{tab:comparison}.

\begin{table}[htpb]
	\centering
	\caption{Comparison of RMSE, MAE and computation time of our model against the benchmarked VQ-VAE and HMM.
		The choice of prior model impacts performances, yet simple architectures are able to model the dependencies of the observations.
		Mean values of the estimators over the validation samples are reported with their variance.}
	\label{tab:comparison}
	\resizebox{0.45\textwidth}{!}{\begin{tabular}{llll}
		\toprule
		ETT dataset                 & RMSE            & MAE             & Computation time \\
		\toprule
		Ours (gru)                  & $0.21 \pm 0.10$ & $0.17 \pm 0.09$ & $39 ms$          \\
		Ours (rnn)                  & $0.21 \pm 0.10$ & $0.17 \pm 0.09$ & $39 ms$          \\
		Ours (cnn)                  & $0.20 \pm 0.10$ & $0.16 \pm 0.09$ & $131 ms$         \\
		VQ-VAE                      & $0.28 \pm 0.12$ & $0.24 \pm 0.12$ & $39 ms$          \\
		HMM                         & $0.44 \pm 0.13$ & $0.32 \pm 0.10$ & $994 ms$         \\
		\toprule
		Building management dataset & RMSE            & MAE             & Computation time \\
		\toprule
		Ours (gru)                  & $0.30 \pm 0.19$ & $0.24 \pm 0.15$ & $39 ms$          \\
		Ours (rnn)                  & $0.30 \pm 0.20$ & $0.25 \pm 0.17$ & $39 ms$          \\
		Ours (cnn)                  & $0.31 \pm 0.20$ & $0.25 \pm 0.16$ & $131 ms$         \\
		VQ-VAE                      & $0.55 \pm 0.33$ & $0.51 \pm 0.32$ & $39 ms$          \\
		HMM                         & $0.46 \pm 0.19$ & $0.40 \pm 0.17$ & $994 ms$         \\
		\bottomrule
	\end{tabular}}
\end{table}

\section{Conclusion}
In this paper, we introduced a generative model with discrete latent variables for time series.
We illustrated our proposed framework using simple prior architectures, which are able to model dependencies in the latent space and allow joint training of all parameters.
However, our methodology can also be adapted for much more complex priors, such as the diffusion bridges models presented in \cite{Cohen2022DB}.

While choosing a suitable number of codebook remains an open question, discrete latents offer new ways of interpreting the model, for instance through regime segmentation.

\clearpage
\bibliographystyle{IEEEtran}
\bibliography{references}
\end{document}